\documentclass{article}

\PassOptionsToPackage{numbers, compress}{natbib}
\usepackage[preprint]{neurips_2021}

\usepackage[utf8]{inputenc} 
\usepackage[T1]{fontenc}    
\usepackage{hyperref}       
\usepackage{url}            
\usepackage{booktabs}       
\usepackage{amsfonts}       
\usepackage{nicefrac}       
\usepackage{microtype}      
\usepackage{xcolor}         
\usepackage{graphicx}
\usepackage{amsmath}
\title{Gaudí: Conversational Interactions with Deep Representations to Generate Image Collections}

\author{%
  Victor S. Bursztyn\\
  Northwestern University\\
  Evanston, IL 60201 \\
  \texttt{v-bursztyn@u.northwstern.edu} \\
  \And
  Jennifer Healey\\
  Adobe Research\\
  San Jose, CA 95110 \\
  \texttt{jehealey@adobe.com} \\
  \And
  Vishwa Vinay\\
  Adobe Research\\
  San Jose, CA 95110 \\
  \texttt{vinay@adobe.com} \\
}

\begin{document}

\maketitle

\section{Introduction}

\begin{figure}[htb]
  \centering
  \includegraphics[width=.9\textwidth]{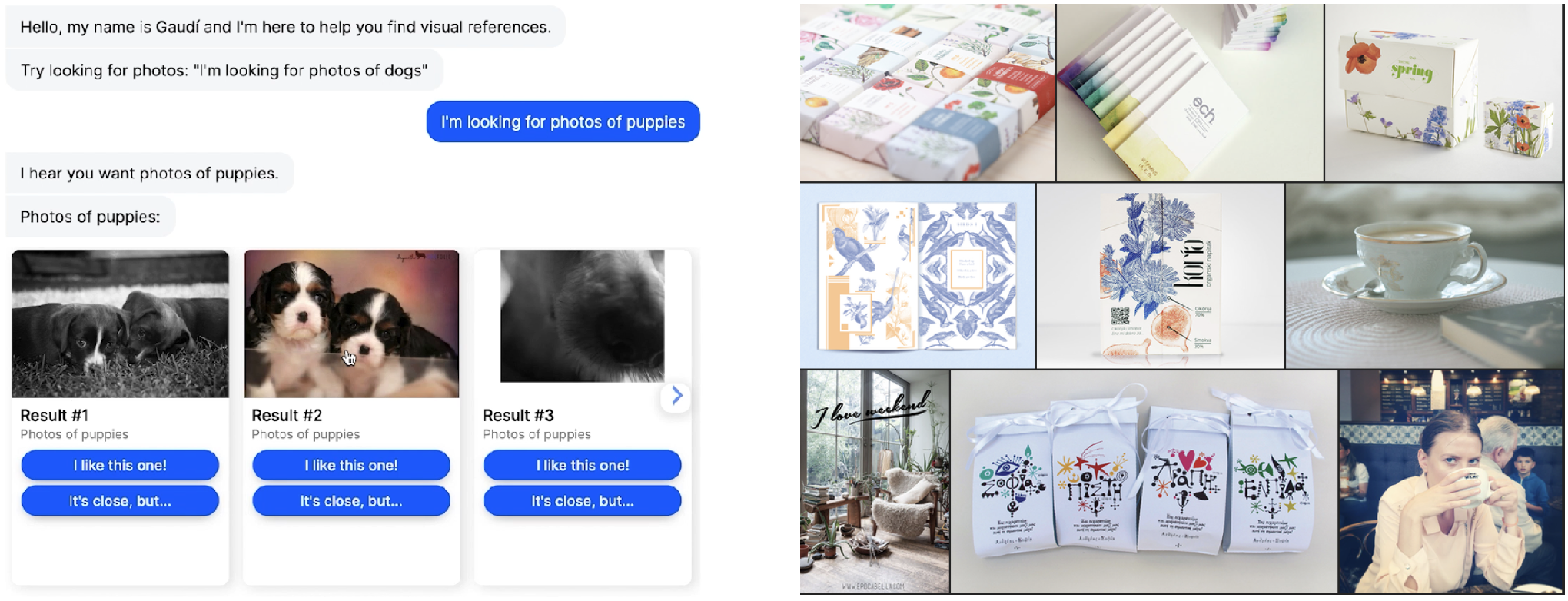}
  \caption{Left: Gaudí responding to a user query ``I'm looking for photos of puppies.'' Right: A mood-board created by a professional designer using Gaudí for the given project briefing: ``You're designing a new ecofriendly, highend coffee brand that is notorious for its floral flavors.'' All images are from the BAM dataset \cite{wilber2017bam}.}
  \label{fig:example}
\end{figure}

Gaudí was developed to help designers search for inspirational images using natural language. In the early stages of the design process, designers will typically create thematic image collections called ``mood-boards'' (example shown in Fig. \ref{fig:example}) in order to elicit and clarify a client's preferred creative direction. Creating a mood-board involves sequential image searches which are currently performed using keywords or images. Gaudí transforms this process into a conversation where the user is gradually detailing the mood-board's theme. This representation allows our AI to generate new search queries from scratch, straight from a project's briefing, following a hypothesized mood.



Previous computational approaches to this process 
tend to oversimplify the decision space, seeking to define it by hard coded qualities like dominant color, saturation and brightness \cite{koch2020imagesense, koch2019may}. Recent advances in realistic language modeling (e.g., with GPT-3 \cite{brown2020language}) and cross-modal image retrieval (e.g., with CLIP \cite{radford2021learning}) now allow us to represent image collections in a much richer semantic space, acknowledging richer variation in the \textit{stories designers tell} when presenting a creative direction to a client.

\section{Methods}

In this section, we present the following methods: image retrieval based on text only; image retrieval based on both a reference image and a text query (or simply “composed image retrieval” \cite{liu2021image}); and mood-board generation by (i) using a project briefing to generate a natural language story, and (ii) using each step of this story as a text query for image retrieval.

\begin{figure}[htb]
  \centering
  \includegraphics[width=.7\textwidth]{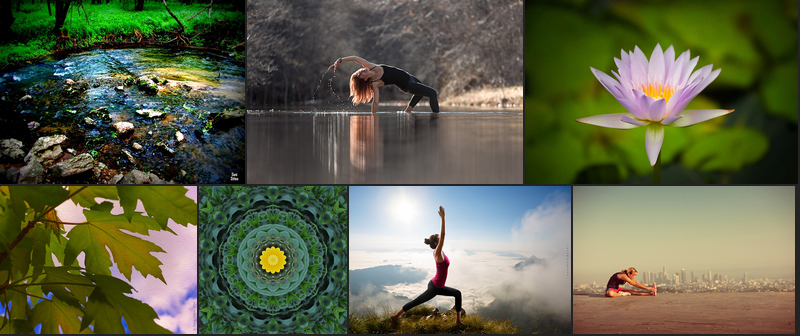}
  \caption{An automatically generated mood-board for the new project briefing: ``You're designing a new yoga kit for a highend company that is famous for its athletic clothes.'' Images from BAM \cite{wilber2017bam}.}
  \label{fig:generated}
\end{figure}

\textbf{Method \#1:} Let $q$ be a text query (e.g., “I'm looking for photos of puppies”) and $\Phi_q$ its cross-modal CLIP embedding. Let $D$ be our image dataset and $i$ an image $i \in D$, then $\Phi_i$ denotes the cross-modal CLIP embedding of $i$. The pairwise similarity between $q$ and $i$ can be denoted by $Sim(q,i) = cos(\Phi_q, \Phi_i)$ such that text-based image retrieval can be defined as $Retr(q,D) = argmax_{i \in D}Sim(q,i)$.

\textbf{Method \#2:} Let $q_m$ be a multi-modal query combining a reference image $r$ (e.g., a previously selected puppy photo) with a text query $q$ (e.g., “I want it more cheerful”) such that $q_m = \Phi_r \oplus \Phi_q$. To account for the extended shape of $q_m$ and due to the cross-modal nature of CLIP, we generate an extended embedding $\Phi_i^{ext}$ for each image $i \in D$ defined as $\Phi_i^{ext} = \Phi_i \oplus \Phi_i$. Again, the pairwise similarity between $q_m$ and $i$ can be denoted by $Sim(q_m,i) = cos(q_m, \Phi_i^{ext})$ such that composed image retrieval can be defined as $Retr(q_m,D) = argmax_{i \in D}Sim(q_m,i)$.

\textbf{Method \#3:} Let $b$ be a project briefing expressed in natural language and $Q$ a set of queries made by a professional designer when creating a mood-board. We condition GPT-3 with one fully detailed example (single-shot) of $b \rightarrow Q$, as described below, plus a new briefing for which we want to generate search queries. We use the DaVinci model with temperature = 0.7, top p = 1.0, response length = 80, and no penalties when sampling the output. Each generated query serves as input to method \#1 or \#2 in order to generate a mood-board as seen in Fig. \ref{fig:generated}.

\tiny
\label{eq:1}
\begin{align*}
&{\rm \bf GPT3\,Input:} \\[-1pt]
&You're\,designing\,a\,new\,ecofriendly,\,highend\,coffee\,brand\,that\,is\,notorious\,for\,its\,floral\,flavors.\,=> \\[-2pt]
&I'm\,looking\,for\,photos\,of\,women\,sipping\,coffee. \
I'm\,looking\,for\,photos\,of\,joyful\,coffee\,packages. \\[-2pt]
&I'm\,looking\,for\,photos\,of\,coffee\,cups\,and\,books. \
I'm\,looking\,for\,photos\,of\,luxury\,coffee\,shops\,with\,plants. \\[-2pt]
&I'm\,looking\,for\,images\,of\,floral\,packaging. \
I'm\,looking\,for\,images\,of\,floral\,packaging\,that\,seems\,a\,bit\,more\,craft. \\[-2pt]
&I'm\,looking\,for\,images\,of\,blue,\,floral\,packaging\,that\,seems\,a\,bit\,more\,craft. \\[-2pt]
&I'm\,looking\,for\,images\,of\,classy,\,colored,\,craft\,packaging. \
I'm\,looking\,for\,images\,of\,posters\,with\,blue\,birds\,and\,flowers. \\[-2pt]
&I'm\,looking\,for\,images\,of\,posters\,with\,blue\,birds\,and\,flowers. \\[-2pt]
&You're\,designing\,a\,new\,yoga\,kit\,for\,a\,highend\,company\,that\,is\,famous\,for\,its\,athletic\,clothes.\,=> \\[-1pt]
&{\rm \bf GPT3\,Output:} \\[-1pt]
&I'm\,looking\,for\,photos\,of\,trees\,and\,grass. \
I'm\,looking\,for\,photos\,of\,water. \\[-2pt]
&I'm\,looking\,for\,photos\,of\,leaves. \
I'm\,looking\,for\,images\,of\,women\,practicing\,yoga. \\[-2pt]
&I'm\,looking\,for\,images\,of\,women\,practicing\,yoga\,and\,wearing\,athletic\,clothes. \\[-2pt]
&I'm\,looking\,for\,images\,of\,women\,practicing\,yoga\,and\,wearing\,athletic\,clothes\,in\,nature. \\[-2pt]
&I'm\,looking\,for\,images\,of\,women\,practicing\,yoga\,in\,nature.
\nonumber
\end{align*}
\normalsize


\section{Results}

We recruited a professional designer for an in-depth exploration of Gaudí, leading to the ground-truth in method \#3 and expert assessments of the automatically generated mood-boards (e.g., Fig. \ref{fig:generated}). The subject was very satisfied (5 in a 5-point scale) with the responsiveness of methods \#1 and \#2 afforded by CLIP. When asked about the quality of the generated mood-boards, the subject rated 4.5 (of 5) for the queries generated by GPT-3 and 3 (of 5) for the mood-boards. Besides a positive surprise with the queries, the subject suggested: ``Although I would have picked different images, I see a story. This may be useful to marketers that use mood-boards in their work but are not used to crafting them.''





{
\small

\bibliography{refs}

}




\end{document}